\documentclass[conference,10pt,letterpaper,twoside]{style/ieeeconf}

\IEEEoverridecommandlockouts    % This command is only
\makeatletter 

\let\proof\@undefined
\let\endproof\@undefined

\let\NAT@parse\undefined
\makeatother

\usepackage[x11names]{xcolor}

\usepackage[ruled, vlined, linesnumbered]{algorithm2e}
\makeatletter
\newcommand{\removelatexerror}{\let\@latex@error\@gobble}
\makeatother
\SetNlSty{bfseries}{\color{black}}{}
\usepackage{xcolor}

\SetCommentSty{mycommfont}

\usepackage[numbers, sort, compress]{natbib}
\usepackage{graphicx}
\usepackage{amsmath,amssymb}
\usepackage{nicematrix}
\usepackage{caption}
\usepackage{subcaption}
\usepackage[export]{adjustbox}

\usepackage{amsthm}  

\usepackage{style/perl_acronyms}
\usepackage{style/perl_math}
\usepackage{style/perl_SIunits}
\usepackage{style/perl_misc}
\usepackage{multirow}
\usepackage{verbatimbox}
\usepackage[normalem]{ulem}

\usepackage{array}
\newcolumntype{C}[1]{>{\centering\arraybackslash}p{#1}}

\usepackage{hyperref}

\usepackage{fancyhdr}
\fancypagestyle{withfooter}{
  
  \fancyhead[L]{}
  \fancyhead[R]{}
  \fancyfoot[C]{\footnotesize Presented at the 2025 IEEE ICRA Workshop on Field Robotics}
}
\usepackage{flushend}

\newcommand{\q}[0]{\quad}

\usepackage[normalem]{ulem}

\begin{document}
\title{Feature Geometry for Stereo Sidescan and Forward-looking Sonar}

\author{Kalin~Norman and Joshua~G.~Mangelson%
    \thanks{This work was funded under Department of Navy awards N00014-21-1-2435 and N00014-21-1-2260 issued by the Office of Naval Research.}
  \thanks{K.~Norman and J.~Mangelson, are at Brigham Young University in Provo, Utah. They can be reached at: \texttt{\{kalinnorman, mangelson\}@byu.edu}}
}

\maketitle
\begin{abstract}

In this paper, we address stereo acoustic data fusion for marine robotics and propose a geometry-based method for projecting observed features from one sonar to another for a cross-modal stereo sonar setup that consists of both a forward-looking and a sidescan sonar.
Our acoustic geometry for sidescan and forward-looking sonar is inspired by the epipolar geometry for stereo cameras, and we leverage relative pose information to project where an observed feature in one sonar image will be found in the image of another sonar.
Additionally, we analyze how both the feature location relative to the sonar and the relative pose between the two sonars impact the projection.
From simulated results, we identify desirable stereo configurations for applications in field robotics like feature correspondence and recovery of the 3D information of the feature.

\end{abstract}

\acresetall

\IEEEpeerreviewmaketitle

\section{Introduction}

\begin{figure}
    \centering
    \includegraphics[width=0.8\linewidth]{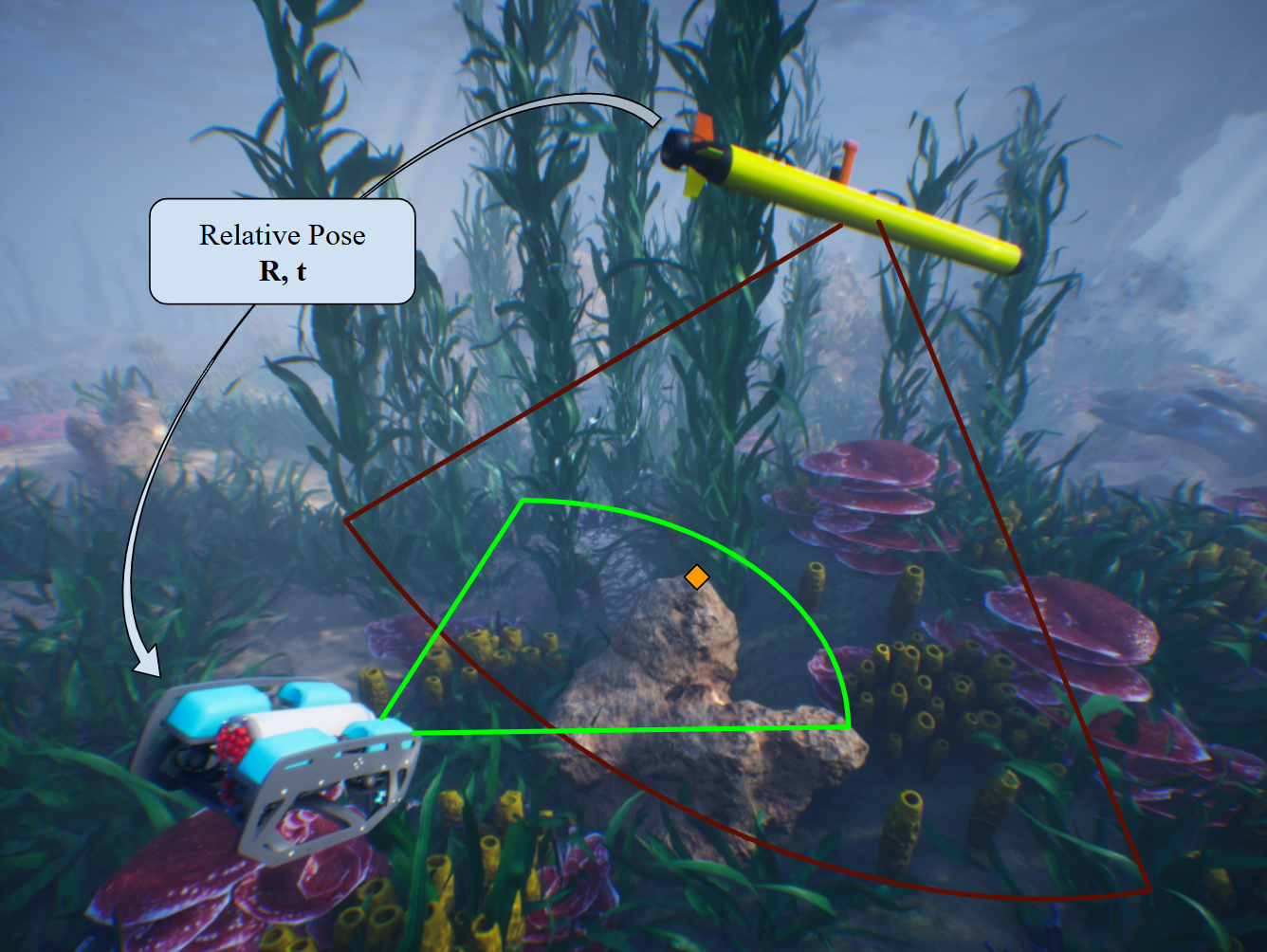}
    \caption{With a known relative pose between two sonar sensors, even with different modalities -- as with forward-looking and sidescan sonar -- we enable more accurate cross-modal data association through a geometric projection.}
    \label{fig:wksp_cm_robots}
\vspace{-15pt}
\end{figure}

Field robotic applications, such as localization and mapping, in underwater environments face significant challenges due to the complex and dynamic nature of the marine domain. 
Unlike terrestrial or aerial settings, water introduces constraints on common sensors, such as GPS and optical cameras, through signal attenuation and variable environmental conditions \cite{tan2011survey, zhou2023underwater}.
Traditional sonar-based sensing methods operate well in marine environments, although they have their own limitations such as low resolution, high noise levels, and sensitivity to environmental disturbances.
To address these challenges, cross-modal acoustic sensor fusion to integrate data from multiple sensor modalities offers a novel approach to overcome some of the limitations of a single acoustic sensor. 
Cross-modal sensor fusion is possible through the use of multiple sensors on a single robot, or via multi-agent operations, as shown in Fig. \ref{fig:wksp_cm_robots}.
This fusion can enable greater resilience to noise and occlusion, as well as improved methods for feature extraction, state estimation, and ultimately advance the capabilities of autonomous underwater systems for exploration, inspection, and mapping.

Optical sensors are nearly ubiquitous in field robotics, though the loss of depth information is a well-known limitation of monocular optical cameras. 
In order to mitigate this limitation, the theory of epipolar geometry has been well studied and enables the restoration of depth from multiple views \cite{lerman1970computer}.
The theory of epipolar geometry has since played a fundamental role in a variety of applications, including localization, mapping, reconstruction, and many others \cite{martinez2014taxonomy}.
Although useful for those applications, optical sensors are poorly suited for underwater use due to turbidity, backscatter, and the absorption of light in water.

In marine environments, acoustic signals propagate more clearly than optical signals \cite{diamant2017relationship}, and as a consequence sonar sensors are more widely used for underwater perception.
Like cameras, sonars project 3D information to a lower-dimensional representation to form an image.
Stereo sonar systems are surprisingly underrepresented in the literature, but the important work of Negahdaripour provides an initial analysis of the geometry of a forward-looking stereo sonar setup used for reconstruction \cite{negahdaripour2018analyzing}.
We believe that stereo sonar systems can help recover lost information and enable algorithms for acoustic sensors that are analogous to the computer vision algorithms that are widely used in field robotics. 

In this paper we build upon prior work to enable cross-modal stereo acoustic systems that incorporate both forward-looking and sidescan sonar by: 
(1) Deriving geometric projections for the cross-modal stereo sonar configurations of a forward-looking to a sidescan sonar, and a sidescan to a forward-looking sonar,
(2) Providing an analysis for each stereo setup to show how the geometric projections can be used to inform both hardware configurations and software algorithms, and
(3) Identifying clear applications where the geometric projections can enable cross-modal improvements to current state of the art systems for robotic localization, mapping, reconstruction, and more.

\section{Related Works}

Computer vision algorithms have used epipolar geometry for decades \cite{lerman1970computer}.
Epipolar geometry for optical sensors constrains the location where a feature in one image might be found in another given a known relative pose between the sensors. 
For many computer vision algorithms, including the recovery of depth information \cite{rajagopalan2004depth}, epipolar geometry is an essential component.
Our work is related to epipolar geometry in that we use the same geometric concepts to determine how a feature visible in one sonar image projects to the image space of a second sonar given a known relative pose between the sensors. 

\begin{figure}[t]
    \centering
    \begin{subfigure}{0.35\textwidth}
        \centering
        \includegraphics[width=\textwidth]{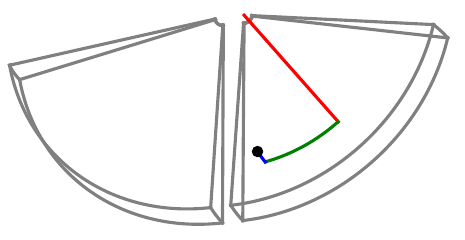}
        \caption{}
        \label{fig:wksp_ssproja}
    \end{subfigure}
    \par\vspace{2pt}
    \begin{subfigure}{0.35\textwidth}
        \centering
        \includegraphics[width=\textwidth]{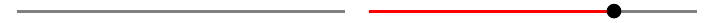}
        \caption{}
        \label{fig:wksp_ssprojb}
    \end{subfigure}
    \par\vspace{2pt}
    \begin{subfigure}{0.35\textwidth}
        \centering
        \includegraphics[width=\textwidth]{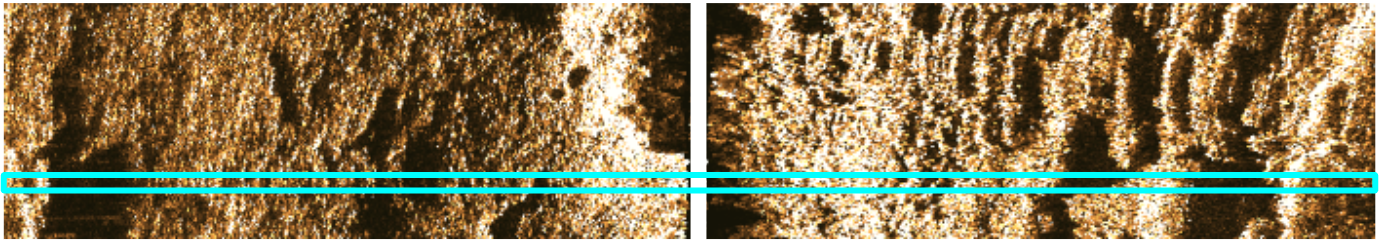}
        \caption{}
        \label{fig:wksp_ssprojc}
    \end{subfigure}
    \caption{Sidescan sonar model. (a) The field of view of a left and right transducer with an observed feature shown as a black dot, with the range, azimuth, and elevation components in red, blue, and green, respectively. (b) Data output from a sidescan sonar, with possible range measurements shown in gray. The range of the feature is shown for visual clarity, in red, with the feature itself as a black dot at the appropriate range. (c) Example of a real sidescan sonar image that consists of a sequence of measurements, one of which is outlined in teal.}
    \label{fig:wksp_ssproj}
\vspace{-15pt}
\end{figure}

\begin{figure}[h]
    \centering
    \begin{subfigure}{0.265\textwidth}
        \centering
        \includegraphics[width=\textwidth]{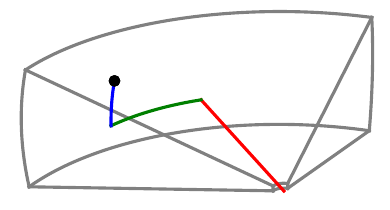}
        \caption{}
        \label{fig:wksp_flproja}
    \end{subfigure}
    \hfill
    \begin{subfigure}{0.205\textwidth}
        \centering
        \includegraphics[width=\textwidth]{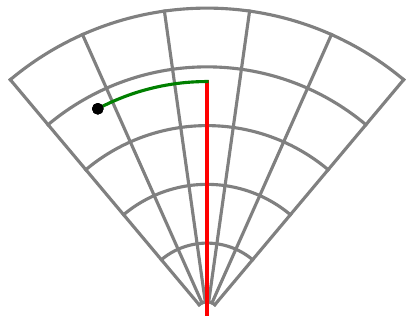}
        \caption{}
        \label{fig:wksp_flprojb}
    \end{subfigure}
    \begin{subfigure}{0.25\textwidth}
        \centering
        \includegraphics[width=\textwidth]{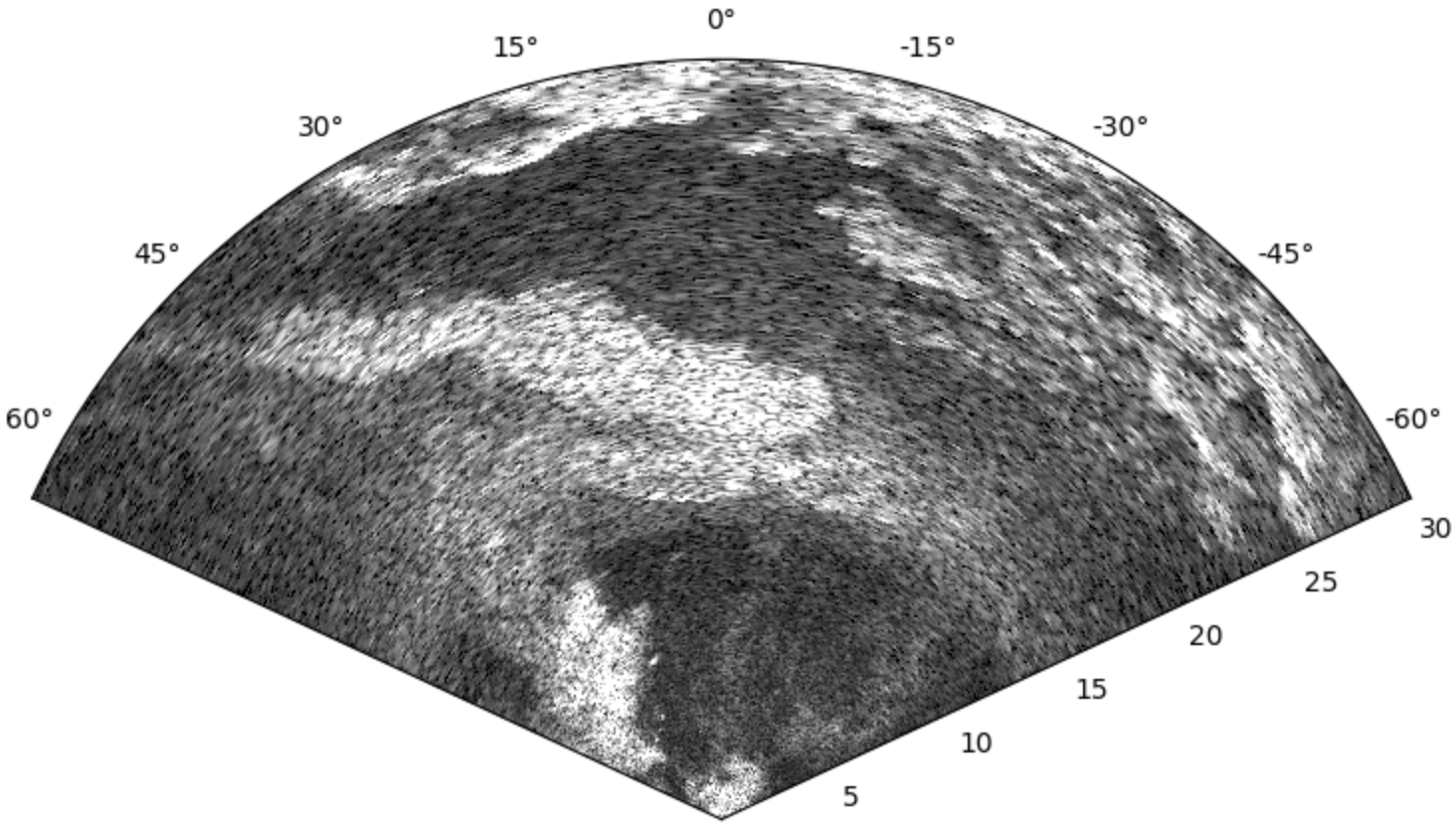}
        \caption{}
        \label{fig:wksp_flprojc}
    \end{subfigure}
    \caption{Forward-looking sonar model. (a) The field of view of the sensor, with a point in 3D space and the range, azimuth and elevation in red, green, and blue, respectively. (b) Forward-looking sonar data output with the range, azimuth, and projected point. (c) Example of a real forward-looking sonar image labeled with range in meters and azimuth in degrees.}
    \label{fig:wksp_flproj}
\vspace{-15pt}
\end{figure}

The efficacy of optical sensors in marine environments is limited due to factors such as turbidity, the high absorption of light in water, and backscatter from floating particulates \cite{dos2015evaluation}.
Despite their limitations, optical sensors are still used in underwater operations.
Within the last decade there has been a significant increase in cross-modal approaches that seek to overcome some of the optical limitations in systems that fuse optical and acoustic data \cite{ferreira2016underwater}.
Although an excellent example of cross-modal sensor fusion, opti-acoustic systems are, to the best of our knowledge, currently limited to rigid setups to ensure proper calibration \cite{hurtos2010integration}.
Other state of the art cross-modal methods \cite{fu2025deep} continue to rely on the geometry of optical sensors, even in the context of acoustic sensor modalities where this geometry is not representative of how acoustics are used for perception.
In an effort to enable more accurate cross-modal acoustic data fusion, in this paper, we present an outline of the geometry that maps feature points from one sonar to another based upon the acoustic projection model of various sonar sensor configurations. 

Stereo sonar systems are relatively uncommon, but Negahdaripour and his collaborators have been pushing the limits of what is possible with a stereo forward-looking sonar setup, including being the first to derive the geometry for their configuration \cite{negahdaripour2018analyzing}.
His work enabled others to explore feature correspondence for identical stereo forward-looking sonars \cite{gode2024sonic}.
Other efforts have attempted to fuse forward-looking and sidescan sonar data, such as in \cite{liu2025mapping}, where they use a forward-looking sonar to fill in the nadir region that is present in sidescan mosaics, although with significant overhead.
Their approach, and every other cross-modal algorithm that we are aware of at this time, requires a complicated and custom approach to enable data association between two different acoustic sensors. 
We seek to simplify that process by deriving the cross-modal geometry for sidescan and forward-looking sonar.

\section{Sonar Projection Models}

In this work we address forward-looking and sidescan sonars. 
Both sensors emit an acoustic pulse and record intensity of the acoustic return from the environment. 
The physical construction of the sensor determines the field of view, which is typically described in spherical coordinates by the minimum and maximum range and the azimuth and elevation aperture.

Sidescan sonar is designed to have a high range and azimuth aperture and a very narrow elevation aperture. 
Intensity is recorded with respect to the time of flight and is mapped to range, using the speed of sound in water. 
Azimuth and elevation information is lost in the projection.
The simplest way to formulate this mathematically is to take a 3D point in spherical coordinates $(r, \theta, \phi)$ and drop both elevation and azimuth and retain only the range component.
Thus the information obtained about a single feature is limited to the range to that feature. As such, we define the the projection model of a sidescan sonar as follows:  
\begin{equation}
    \Pi_{ss}(r, \theta, \phi) = r
\label{eq:wksp_ss_proj}
\end{equation}
where the subscript of $ss$ indicates that it is for a sidescan sonar. 
The very narrow elevation aperture of sidescan sonar often leads to the assumption that the elevation uncertainty is negligible, but in this paper we do not use that assumption.
A visual representation of the projection model for sidescan sonar is shown in Fig. \ref{fig:wksp_ssproj}.
The most common application of sidescan sonar is for it to either be towed behind a ship or mounted to an autonomous vehicle, facing down towards the seafloor. 
The elevation is inline with the direction of travel so the vehicle collects data about what is below and to the left and right. 

Forward-looking sonar is more similar to an optical camera and is sometimes referred to as an acoustic camera, or imaging sonar. 
When compared to a sidescan sonar, the maximum range and azimuth aperture are usually smaller than that of a sidescan sonar and the elevation aperture is much larger. 
Forward-looking sonar is a type of multi-beam sonar, where beam forming is used to determine the azimuth angle of an acoustic return in addition to the range calculation derived from the time of flight. 
Elevation information is lost, but enough data is recorded to compose a two-dimensional image from each acoustic ping. 
Thus, 3D points are projected from spherical coordinates onto a polar (range and azimuth) plane and we can define the projection model for a forward-looking sonar as follows:
\begin{equation}
    \Pi_{fl}(r, \theta, \phi) = \begin{bmatrix}
        r \\
        \theta
    \end{bmatrix}
\label{eq:wksp_fl_proj}
\end{equation}
where the subscript of $fl$ indicates that it is for a forward-looking sonar. 
The data is most easily visualized as a polar plot where each pixel shows the intensity of the acoustic return for every recorded range and azimuth. 
A visual of the projection model is shown in Fig. \ref{fig:wksp_flproj}.
 
\section{Stereo Sonar Geometry}

\begin{figure}[t]
    \centering
    \begin{subfigure}{0.17\textwidth}
        \centering
        \includegraphics[width=\textwidth]{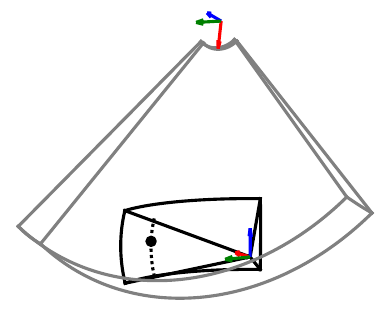}
        \caption{}
        \label{fig:wksp_fl_to_ss_a}
    \end{subfigure}
    \hfill
    \begin{subfigure}{0.17\textwidth}
        \centering
        \includegraphics[width=\textwidth]{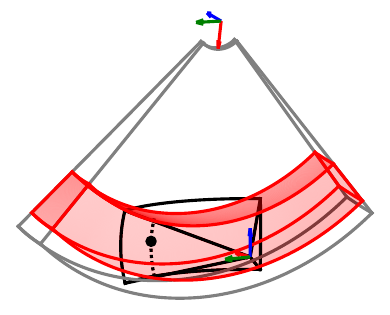}
        \caption{}
        \label{fig:wksp_fl_to_ss_b}
    \end{subfigure}
    \hfill
    \begin{subfigure}{0.13\textwidth}
        \centering
        \includegraphics[width=\textwidth, angle=270]{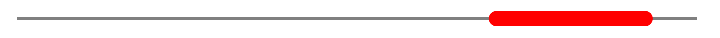}
        \caption{}
        \label{fig:wksp_fl_to_ss_c}
    \end{subfigure}
    \caption{Feature projection from forward-looking to sidescan sonar: (a) The forward-looking sonar faces into the page and observes a feature, represented by a black dot, and the corresponding arc in 3D space, given by the known elevation aperture, is shown by a dotted line. (b) The sidescan sonar can observe the same point and given the relative pose between the sensors, the arc is mapped to the range of the sidescan sonar, which results in a volume has spans part of the range within the sensor field of view, as well as the azimuth and elevation apertures, with the volume shown in red. (c) The 3D volume is projected to the measurement space of the sidescan sonar, shown in red, and is the region in which the feature can be found.}
    \label{fig:wksp_fl_to_ss}
\vspace{-15pt}
\end{figure}

Given a stereo sonar configuration, we consider the projection from forward-looking to sidescan sonar, and from sidescan to forward-looking sonar.
For each scenario, we assume that a feature is observed in the data from the first sonar.
Given a known relative pose and the assumption that there is no noise or uncertainty in the measurement or the relative pose, our goal is to determine the subset of the data space of the second sonar to which the feature might project.

The general process is to: (1) identify the feature, (2) use the inverse projection model of the first sonar to determine a region or set of possible three-dimensional location of the feature, (3) change reference frames such that the space of possible locations is now determined relative to the second sonar, (4) limit the region to the portion of 3D space that can be seen by the second sonar, and (5) use the projection model for the second sonar to obtain the result.

\subsubsection{Feature Detection}

The first step is to uniquely identify a feature point as observed in the sonar data. 
For sidescan sonar, this would be carried out by determining the range at which a specific point of interest is found. 
In forward-looking sonar, the range and azimuthal angle of a feature point would be determined.

\subsubsection{Back-Projection into 3D}

Once the low degree of freedom observations of the feature have been determined, we then back-project the feature point into 3D space to determine a set of 3D points that are possible locations of the observed feature. 
The details of this process are sensor modality specific and will be presented in more detail in the next two subsections. The result of this step is a set of 3D points, represented in spherical coordinates with respect to the reference frame of the first sonar, where any given point in the set can be represented as $\mathbf{p}_s = \begin{bmatrix}
    r & \theta & \phi
\end{bmatrix}^T$, where we use a subscript of $s$ to denote that the point is in spherical coordinates. 

\subsubsection{Transforming the Sonar Reference Frame}

The next step is to transform the possible 3D locations of our feature to the reference frame of the second sonar. 
To do so we must first convert to Cartesian coordinates 
\begin{equation}
    \mathbf{p}_c = \begin{bmatrix}
        x \\
        y \\
        z
    \end{bmatrix} = \begin{bmatrix}
        r \cos \theta \cos \phi \\
        r \sin \theta \cos \phi \\
        r \sin \phi
    \end{bmatrix}
    \label{eq:wksp_ps1_to_pc1}
\end{equation}
where the subscript $c$ denotes that the point is in Cartesian coordinates.
We can then transform the point to the reference frame of the second sonar using the known rotation matrix, $\mathbf{R}$, and translation vector, $\mathbf{t}$, to get
\begin{equation}
    \mathbf{p}'_c = \mathbf{R} \mathbf{p}_c + \mathbf{t}
    \label{eq:wksp_trans_simp}
\end{equation}
where $'$ is used to indicate that $\mathbf{p}$ is defined with respect to the reference frame of the second sonar.

\subsubsection{Clipping to the Sonar Field of View}

Once the point is defined with respect to the second sonar frame, we need to clip any possible 3D locations that lie outside the second sonar field of view. 
We do this by converting back to spherical coordinates to obtain
\begin{equation}
    \mathbf{p}'_s = \begin{bmatrix}
        r ' \\
        \theta ' \\
        \phi '
    \end{bmatrix} = \begin{bmatrix}
        \sqrt{ (x ')^2 + (y ')^2 + (z ')^2 } \\
        \tan^{-1} \left( \frac{ y ' }{ x ' } \right) \\
        \sin^{-1} \left( \frac{ z ' }{ \sqrt{ (x ')^2 + (y ')^2 + (z ')^2 } } \right)
    \end{bmatrix}.
    \label{eq:wksp_pc2_to_ps2}
\end{equation}
We then clip any possible 3D locations that fall outside the field of view by checking to see if the 3D points fall within the valid minimum and maximum range, and azimuth and elevation apertures of the sensor.
The set of valid 3D points is given by $\mathbf{V}$, where we refer to a point in the set with $\mathbf{v}'_s = \begin{bmatrix}
    r' & \theta' & \phi'
\end{bmatrix}^T$.

\subsubsection{Final Sonar Projection}

Finally, we project $\mathbf{V}$ to the sonar image (for forward-looking sonar) or range axis (for sidescan sonar) according to the corresponding projection method appropriate for the sensor modality. 

When $\mathbf{V}$ is not known, which is usually the case, it is possible to apply a single equation to get $\mathbf{p}'_s$, after which it is necessary to determine $\mathbf{V}$ before performing the final sonar projection.
If $\mathbf{V}$ is known, it is possible to apply a single equation to carry out the full projection process. 
The expanded equations for the general case when $\mathbf{V}$ is unknown, as well as for when it is known, are shown in Table \ref{tab:wksp_proj_eqns}, given that
\begin{equation}
    \mathbf{R} = \begin{bmatrix}
        R_{11} & R_{12} & R_{13} \\
        R_{21} & R_{22} & R_{23} \\
        R_{31} & R_{32} & R_{33} 
    \end{bmatrix} ~~ \text{and} ~~  \mathbf{t} = \begin{bmatrix}
        t_1 \\
        t_2 \\
        t_3
    \end{bmatrix}.
\end{equation}

\begin{table*}[]
    \centering
    \begin{tabular}{||c|c||}
        \hline
        Scenario & Equation \\
        \hline
        \hline
        General & $\mathbf{p}'_s = \begin{bmatrix}
            \left( \sum_{i=1}^3 \left( \left( R_{i1} r \cos \theta \cos \phi + R_{i2} r \sin \theta \cos \phi + R_{i3} r \sin \phi + t_i \right)^2 \right) \right)^{\frac{1}{2}} \\
            \tan^{-1} \left( \frac{ R_{21} r \cos \theta \cos \phi + R_{22} r \sin \theta \cos \phi + R_{23} r \sin \phi + t_2 } { R_{11} r \cos \theta \cos \phi + R_{12} r \sin \theta \cos \phi + R_{13} r \sin \phi + t_1 } \right) \\
            \sin^{-1} \left( \frac{ R_{31} r \cos \theta \cos \phi + R_{32} r \sin \theta \cos \phi + R_{33} r \sin \phi + t_3 } { \left( \sum_{i=1}^3 \left( \left( R_{i1} r \cos \theta \cos \phi + R_{i2} r \sin \theta \cos \phi + R_{i3} r \sin \phi + t_i \right)^2 \right) \right)^{\frac{1}{2}} } \right) 
        \end{bmatrix}$ \\
        \hline
        Forward-looking to Sidescan & $d_{ss} (\phi) = \left( \sum_{i=1}^3 \left( \left( R_{i1} r \cos \theta \cos \phi + R_{i2} r \sin \theta \cos \phi + R_{i3} r \sin \phi + t_i \right)^2 \right) \right)^{\frac{1}{2}}$ \\
        \hline
        Sidescan to Forward-looking & $\mathbf{d}_{fl} (\theta, \phi) = \begin{bmatrix}
            \left( \sum_{i=1}^3 \left( \left( R_{i1} r \cos \theta \cos \phi + R_{i2} r \sin \theta \cos \phi + R_{i3} r \sin \phi + t_i \right)^2 \right) \right)^{\frac{1}{2}} \\
            \tan^{-1} \left( \frac{ R_{21} r \cos \theta \cos \phi + R_{22} r \sin \theta \cos \phi + R_{23} r \sin \phi + t_2 } { R_{11} r \cos \theta \cos \phi + R_{12} r \sin \theta \cos \phi + R_{13} r \sin \phi + t_1 } \right) 
        \end{bmatrix}$ \\
        \hline
    \end{tabular}
    \caption{Expanded equations for projection of a point from one sonar to another. (top) General case of transforming a spherical point from the reference frame of the first sonar to the spherical coordinate in the reference frame of the second sonar. (middle) Direct projection from forward-looking to sidescan data, which is only valid when the point is within the field of view of the forward-looking sonar. (bottom) Direct projection from sidescan to forward-looking sonar data, which is only valid when the point is within the field of view of the sidescan sonar.}
    \label{tab:wksp_proj_eqns}
\vspace{-10pt}
\end{table*}

\subsection{Forward-looking to Sidescan Sonar}

When the feature is first identified in a forward-looking sonar, we have 
\begin{equation}
    \mathbf{q} = \begin{bmatrix}
        r \\
        \theta
    \end{bmatrix}.
\end{equation}
To identify the feature region in 3D space, we use the known information from $\mathbf{q}$ and the minimum and maximum elevation that correspond to the elevation aperture, such that $\phi \in [\phi_{\text{min}}, \phi_{\text{max}}]$.
These details define an arc as outlined below:
\begin{equation}
    \mathbf{p}_s ( \phi ) = \begin{bmatrix}
        r \\
        \theta \\
        \phi
    \end{bmatrix}, \q \forall \phi \in [\phi_{\text{min}}, \phi_{\text{max}}].
\end{equation}
An example of such an arc is shown in Fig. \ref{fig:wksp_fl_to_ss_a}.

\begin{figure}[t]
    \centering
    \begin{subfigure}{0.17\textwidth}
        \centering
        \includegraphics[width=\textwidth]{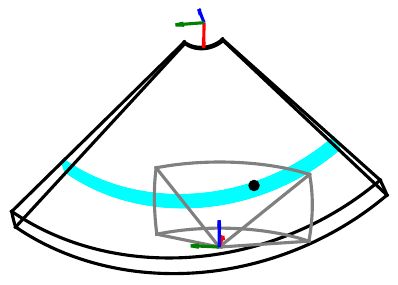}
        \caption{}
        \label{fig:wksp_ss_to_fl_a}
    \end{subfigure}
    \hfill
    \begin{subfigure}{0.17\textwidth}
        \centering
        \includegraphics[width=\textwidth]{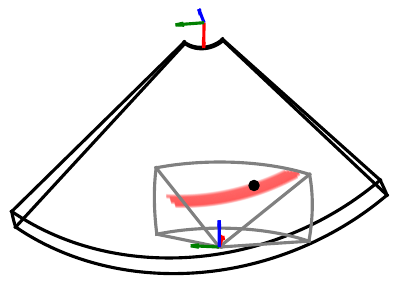}
        \caption{}
        \label{fig:wksp_ss_to_fl_b}
    \end{subfigure}
    \hfill
    \begin{subfigure}{0.13\textwidth}
        \centering
        \includegraphics[width=\textwidth]{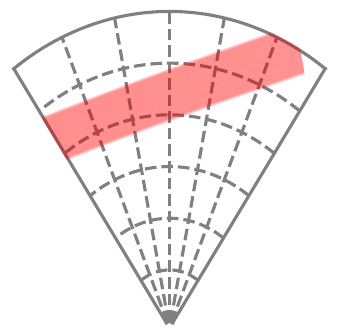}
        \caption{}
        \label{fig:wksp_ss_to_fl_c}
    \end{subfigure}
    \caption{Feature projection from sidescan to forward-looking sonar: (a) The downward facing sidescan sonar observes a feature, represented by a black dot, and the corresponding surface given the known azimuth and elevation apertures is shown in cyan. (b) The forward-looking sonar can observe the same point and given the relative pose between the sensors, the surface is clipped to the portion that is within the forward-looking sonar field of view, where what remains of the original surface is shown in red. (c) The 3D surface is projected to the image space of the forward-looking sonar, shown in red, and is the region in which the feature can be found.}
    \label{fig:wksp_ss_to_fl}
\vspace{-15pt}
\end{figure}

We then transform $\mathbf{p}_s ( \phi )$ to $\mathbf{p}'_s ( \phi )$ and determine $\mathbf{V} (\phi)$, which is a set that consists of points $\mathbf{v}'_s (\phi)$ that form a volume within the field of view of the sidescan sonar.
The final result is obtained through the use of the sidescan sonar projection (eq. \ref{eq:wksp_ss_proj}), applied to the points in $\mathbf{V} (\phi)$ to obtain
\begin{equation}
    \Pi_{ss}(\mathbf{v}'_s (\phi) ) = \Pi_{ss} \left( \begin{bmatrix}
        r'_s ( \phi ) \\
        \theta'_s ( \phi ) \\
        \phi'_s ( \phi )
    \end{bmatrix} \right) = r'_s ( \phi ) .
\end{equation}
A visual example of the projection from a feature identified in a forward-looking sonar image to a measurement from a sidescan sonar is shown in Fig. \ref{fig:wksp_fl_to_ss}.

\subsection{Sidescan to Forward-looking Sonar}

While forward-looking sonar results in an arc for the feature region, sidescan sonar results in a surface. 
Given a sidescan sonar feature measurement, we have
\begin{equation}
    q = r
\end{equation}
as well as the azimuth and elevation apertures of the sensor: $\theta \in [\theta_{\text{min}}, \theta_{\text{max}}]$ and $\phi \in [\phi_{\text{min}}, \phi_{\text{max}}]$.
Consequently, the set of possible 3D locations for an observed feature is defined in spherical coordinates as
\begin{equation}
    \mathbf{p}_s (\theta, \phi) = \begin{bmatrix}
        r \\
        \theta \\
        \phi
    \end{bmatrix}, \q \forall \theta, \phi \in [\theta_{\text{min}}, \theta_{\text{max}}], [\phi_{\text{min}}, \phi_{\text{max}}].
\end{equation}

From $\mathbf{p}_s (\theta, \phi)$ we follow the general process to obtain $\mathbf{p}'_s (\theta, \phi)$ and determine $\mathbf{V} ( \theta, \phi )$, which is a subset of the original surface that lies within the field of view of the forward-looking sonar.
The final result is obtained through the use of the forward-looking sonar projection (eq. \ref{eq:wksp_fl_proj}), applied to each point $\mathbf{v}'_s ( \theta, \phi ) \in \mathbf{V}$ as
\begin{equation}
    \Pi_{fl}(\mathbf{v}'_s ( \theta, \phi ) ) = \Pi_{fl} \left( \begin{bmatrix}
        r'_s ( \theta, \phi ) \\
        \theta'_s ( \theta, \phi ) \\
        \phi'_s ( \theta, \phi )
    \end{bmatrix} \right) = \begin{bmatrix}
        r'_s ( \theta, \phi ) \\
        \theta'_s ( \theta, \phi )
    \end{bmatrix} .
\end{equation}
An example for when a feature is identified in a sidescan sonar measurement and then projected to a forward-looking sonar image is shown in Fig. \ref{fig:wksp_ss_to_fl}.

\section{Application and Analysis}

One of the main applications for the epipolar geometry in stereo vision is feature correspondence. 
Within sonar imagery this is a particularly challenging problem, but the initial work of Negahdaripour \cite{negahdaripour2018analyzing} was used by \cite{gode2024sonic} to train a neural network for feature-matching between stereo forward-looking sonar views. 
A major benefit of this method is that they were able to train on synthetic data, but apply it to real-world data. 
By defining the stereo acoustic geometry for forward-looking and sidescan sonar, we enable similar methods to determine accurate feature correspondence for cross-modal data. 
In addition, these tools will enable enhancement of many of the existing underwater localization and mapping algorithms that rely upon acoustic data, but currently assume projection models defined for optical cameras \cite{yang2024geometry}. 

Moreover, it will also be possible in some cases to determine the elevation information that is lost in the forward-looking sonar projection or the azimuth and elevation information that is lost in the sidescan sonar projection.
Recovery of the lost information is possible when the geometric projection region is large and a correct feature correspondence is determined.
The 3D position of the observed feature is retrieved from the intersection of the known feature location in the data of the first sonar and the inverse geometric projection of the feature from the second sonar back to the first sonar.

As an initial analysis, we evaluate multiple scenarios in simulation, with the intent to inform both hardware configurations and software algorithms.
Our analysis is unable to demonstrate every possibility, but we are able to explore how a variable distance between each sonar and the feature, as well as the relative pose between the sonars, impacts the size of the projection space. The sonar parameters used for the simulation are shown in Table \ref{tab:wksp_sim_params}.

\begin{table}[]
    \centering
    \begin{NiceTabular}{|c|c|c|}
        \hline
        Sensor & Parameter & Value \\
        % \Block{1-2}{Parameter} & & Value (Units) \\ 
        \hline
        \hline
        \Block{4-1}{\rotate Forward-\\looking} & min range & 0.1 (m) \\
        \cline{2-3}
        & max range & 10 (m) \\
        \cline{2-3}
        & azimuth aperture & 60 (deg) \\
        \cline{2-3}
        & elevation aperture & 12 (deg) \\
        \hline
        \Block{4-1}{\rotate Sidescan} & min range & 0.1 (m) \\
        \cline{2-3}
        & max range & 30 (m) \\
        \cline{2-3}
        & azimuth aperture & 130 (deg) \\
        \cline{2-3}
        & elevation aperture & 0.3 (deg) \\
        \hline
    \end{NiceTabular}
    \caption{Sonar parameters for simulation.}
    \label{tab:wksp_sim_params}
\vspace{-15pt}
\end{table}

\begin{figure*}[t]
    \centering
    \includegraphics[width=0.9\linewidth]{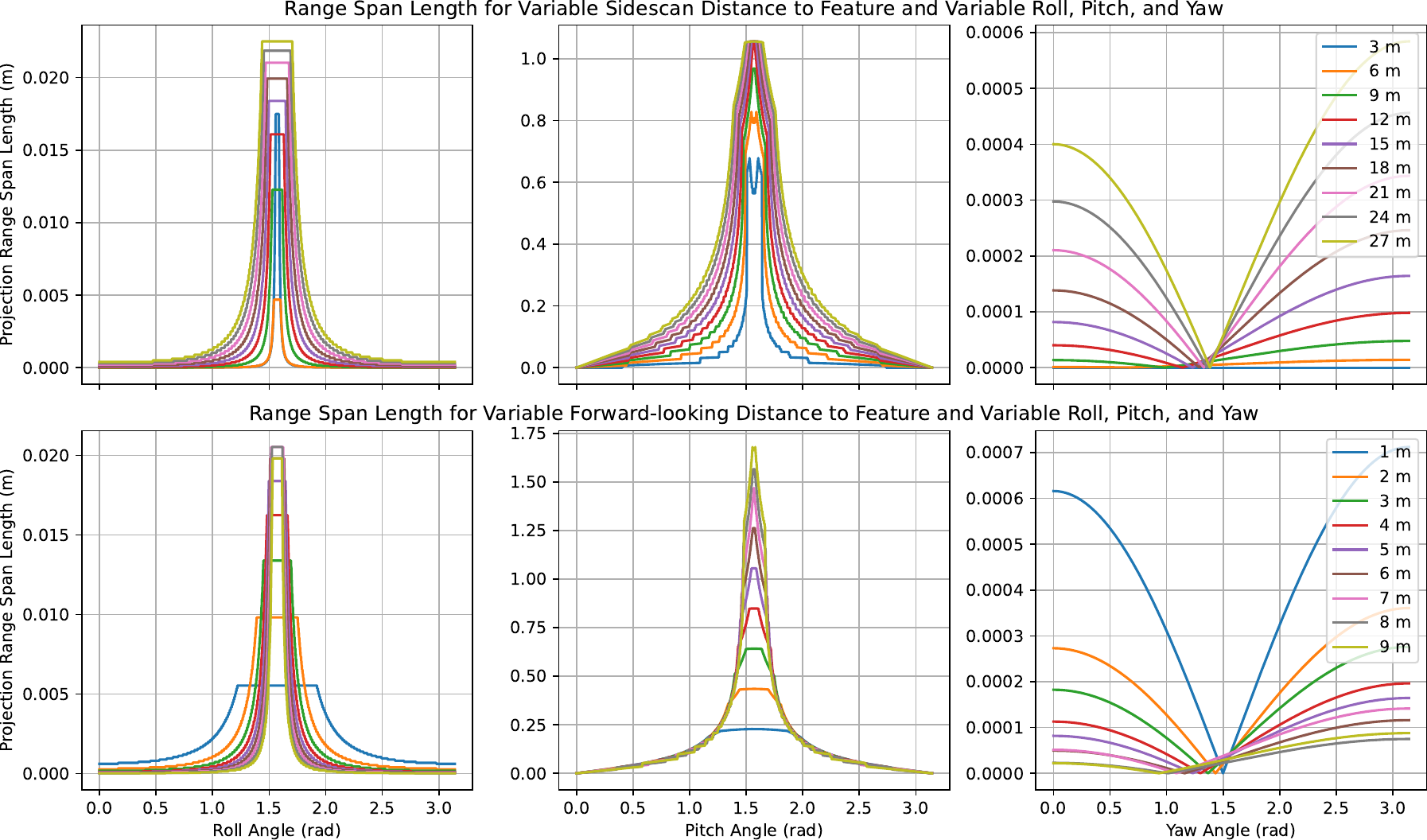}
    \caption{The effects of feature position and isolated relative rotations on the length of the range span. (top row) Results for sweeping the (top left) roll, (top middle) pitch, and (top right) yaw angles, for when the forward-looking sonar is kept at a fixed distance from the feature at 50\% of the maximum range of the sensor. The distance from the sidescan sonar to the feature is varied between 10\% and 90\% of the maximum range of the sensor at 10\% increments. (bottom row) Results for sweeping the (bottom left) roll, (bottom middle) pitch, and (bottom right) yaw angles, for when the forward-looking sonar distance to the feature is varied between 10\% and 90\% of the maximum range of the sensor at 10\% increments. The distance from the sidescan sonar to the feature is fixed at 50\% of the maximum range.}
    \label{fig:wksp_analysis_fl2ss1}
\vspace{-5pt}
\end{figure*}

\begin{figure*}[t]
    \centering
    \includegraphics[width=0.9\linewidth]{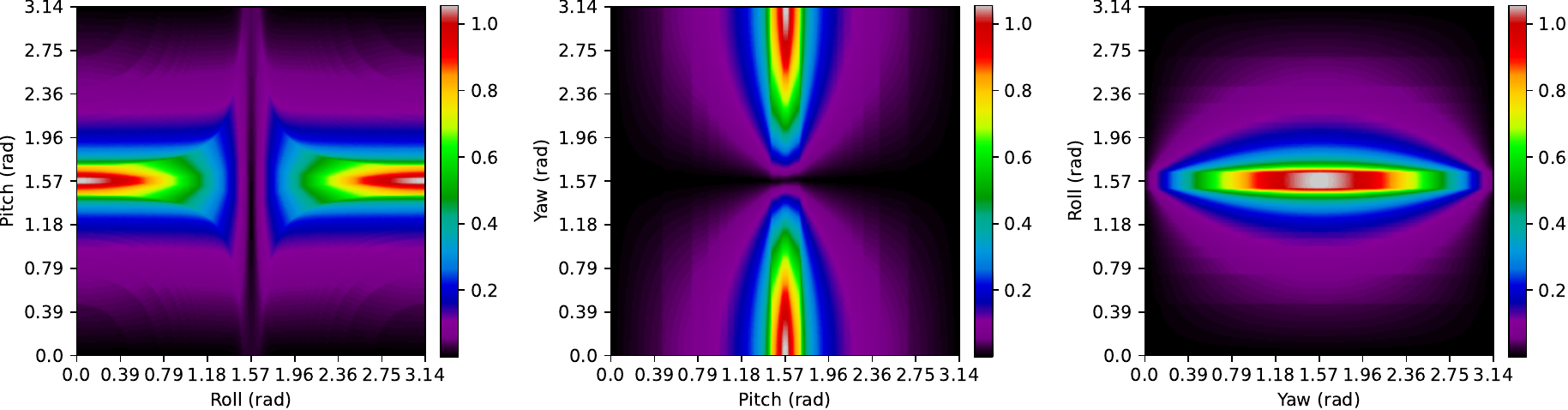}
    \caption{Length of the range span, in meters, as a function of two relative rotation angles between the sensors. For all three rotation combinations the feature is fixed halfway between the minimum and maximum range for both the forward-looking and sidescan sonar. (left) Range span results for sweeping through all combinations of roll and pitch. (middle) Range span results for sweeping through all combinations of pitch and yaw. (right) Range span results for sweeping through all combinations of yaw and roll.}
    \label{fig:wksp_analysis_fl2ss2}
\vspace{-15pt}
\end{figure*}

\subsection{Simulation of Forward-looking to Sidescan Sonar}

For the results presented in this section, we conduct sweeps over the distance from each sonar to the feature, as well as sweeps through the relative roll, pitch, and yaw between the sensors. The feature was always kept at an azimuth and elevation angle of $0$ within the field of view of both sonars, and the translation between the two sensors was computed based off of the desired feature locations for each sensor and relative rotation through
\begin{equation}
    \mathbf{t} = \mathbf{p}_s - \mathbf{R} \mathbf{p}'_s .
\end{equation}
Fig. \ref{fig:wksp_analysis_fl2ss1} shows the results of six different simulations:
(1) The roll angle was swept from $0$ to $\pi$ while the distance from the forward-looking sonar to the feature was fixed at $50\%$ of the maximum range of the sensor and the distance from the sidescan sonar to the feature was varied between $10\%$ and $90\%$ of the maximum range of the sensor in $10\%$ increments. 
(2) Swept the pitch angle from $0$ to $\pi$ with the same feature distance configurations as in the first simulation.
(3) Swept the yaw angle from $0$ to $\pi$ with the same feature distance configurations as in the first simulation.
(4) The roll angle was swept from $0$ to $\pi$ while the distance from the forward-looking sonar to the feature was varied between $10\%$ and $90\%$ of the maximum range of the sensor in $10\%$ increments and the distance from the sidescan sonar to the feature was fixed at $50\%$ of the maximum range of the sensor. 
(5) Swept the pitch angle from $0$ to $\pi$ with the same feature distance configurations as in the fourth simulation.
(6) Swept the yaw angle from $0$ to $\pi$ with the same feature distance configurations as in the fourth simulation.
In all cases, any values not explicitly mentioned were set to a default of $0$.

For Fig. \ref{fig:wksp_analysis_fl2ss2} three additional simulations were performed, so there are nine total simulation scenarios, with tha last three being:
(7) The relative roll and pitch angles were swept between $0$ and $\pi$, where all possible combinations of rotations were included, with the distance from both the forward-looking sonar and the sidescan sonar to the feature fixed at $50\%$ of their respective maximum ranges.
(8) Swept both the relative pitch and yaw angles between $0$ and $\pi$, with all possible combinations of the two rotation angles included, and the distance to the feature for both sensors identical to the first scenario.
(9) Swept both the relative yaw and roll angles between $0$ and $\pi$, with all possible combinations of the two rotation angles included, and the distance to the feature for both sensors identical to the first scenario.
Once again, any values not explicitly mentioned were set to a default of $0$.

\begin{figure*}[h]
    \centering
    \includegraphics[width=0.9\linewidth]{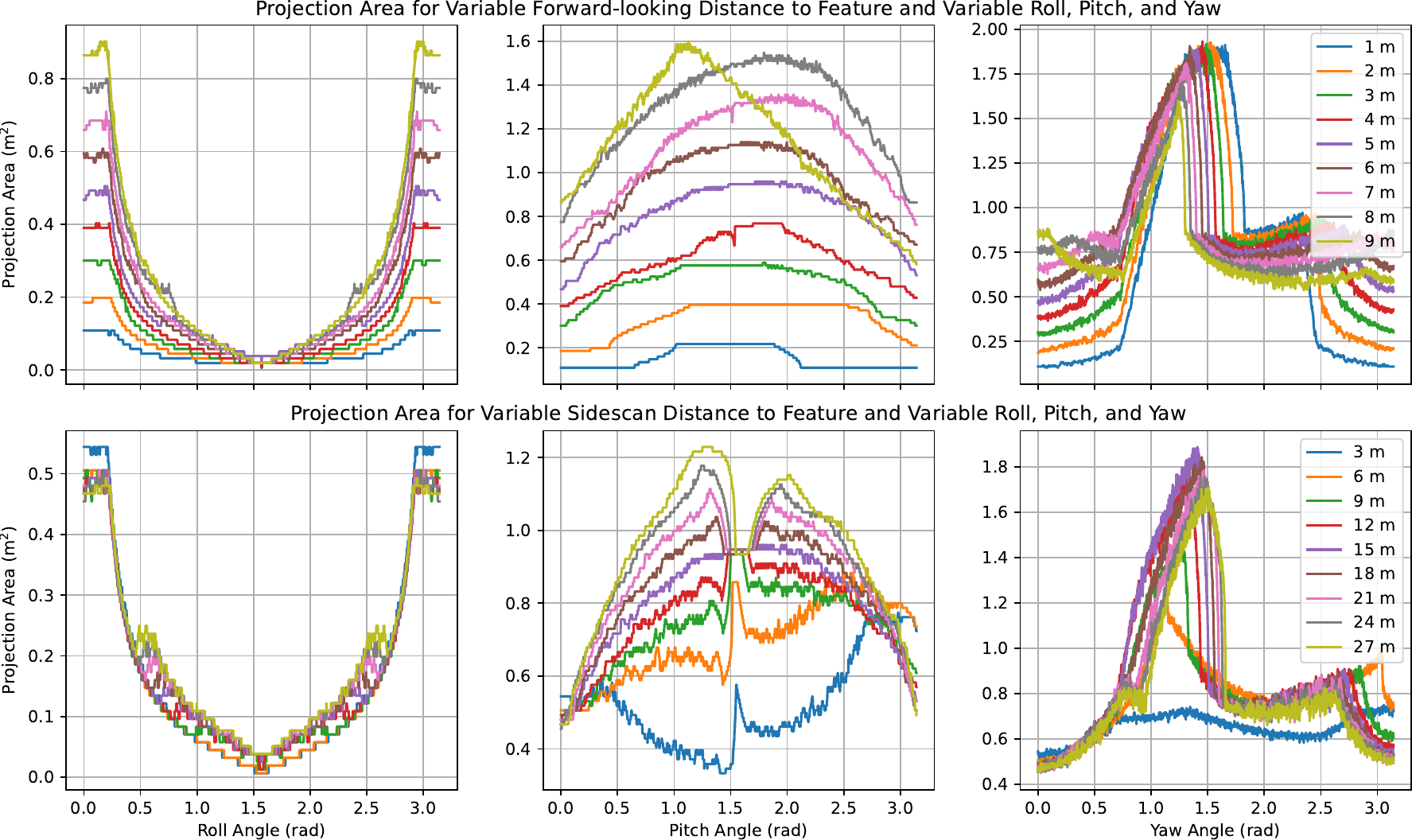}
    \caption{The effects of feature position and isolated relative rotations on the projection area within the forward-looking sonar image. (top row) Results for sweeping the (top left) roll, (top middle) pitch, and (top right) yaw angles, for when the sidescan sonar is kept at a fixed distance from the feature at 50\% of the maximum range of the sensor. The distance from the forward-looking sonar to the feature is varied between 10\% and 90\% of the maximum range of the sensor at 10\% increments. (bottom row) Results for sweeping the (bottom left) roll, (bottom middle) pitch, and (bottom right) yaw angles, for when the sidescan sonar distance to the feature is varied between 10\% and 90\% of the maximum range of the sensor at 10\% increments. The distance from the forward-looking sonar to the feature is fixed at 50\% of the maximum range.}
    \label{fig:wksp_analysis_ss2fl1}
\vspace{-5pt}
\end{figure*}

\begin{figure*}[h]
    \centering
    \includegraphics[width=0.9\linewidth]{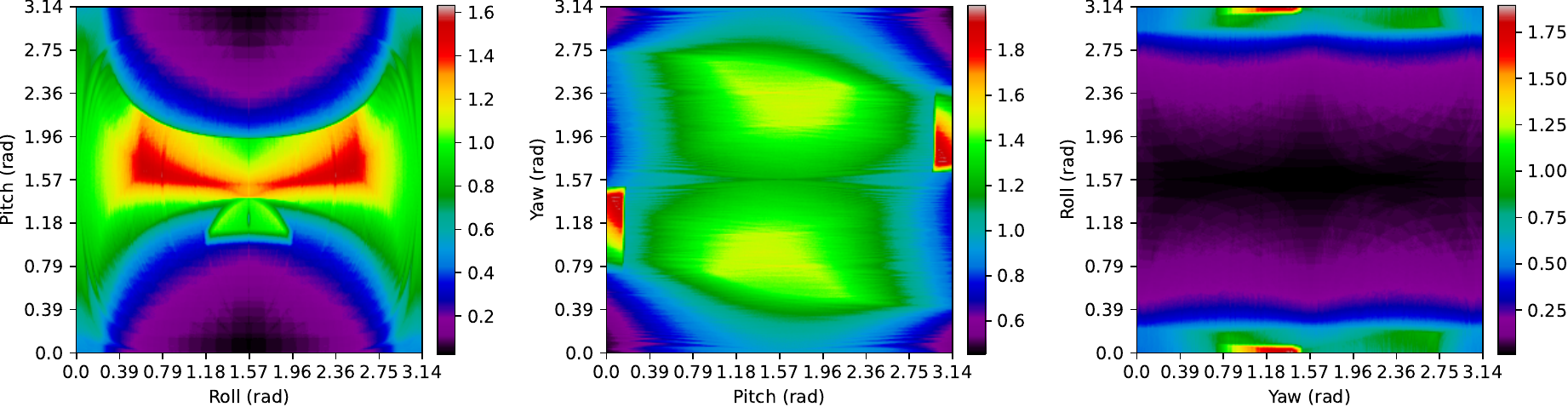}
    \caption{Projection area within the forward-looking sonar image, in meters squared, as a function of two relative rotation angles between the sensors. For all three rotation combinations the feature is fixed halfway between the minimum and maximum range for both the forward-looking and sidescan sonar. (left) Range span results for sweeping through all combinations of roll and pitch. (middle) Range span results for sweeping through all combinations of pitch and yaw. (right) Range span results for sweeping through all combinations of yaw and roll.}
    \label{fig:wksp_analysis_ss2fl2}
\vspace{-15pt}
\end{figure*}

Using the results from our simulations, the two main applications that we consider are feature correspondence and recovering dimensional information that is lost through the individual sensor projections.
For a feature that is identified in a forward-looking sonar image and projected to a sidescan sonar we calculate the length of the span of ranges in the sidescan data (the length of the red segment in Fig. \ref{fig:wksp_fl_to_ss_b}).
The length of the range span is our primary metric in analysis to inform useful configurations for the two applications mentioned previously.
Results for various permutations of $\mathbf{p}_s$, $\mathbf{p}'_s$, and $\mathbf{R}$ are shown in Fig. \ref{fig:wksp_analysis_fl2ss1} and \ref{fig:wksp_analysis_fl2ss2}.

Without picking a specific objective, there is not a relative orientation or feature location that is optimal, as different applications have competing interests. 
Additionally, the sonar parameters and field of view of the sensor used have a significant influence on the resulting analysis. As such, our presented results are specific to our configuration.
However, the overall process is general and can be applied to analyze sensors with any set of configuration parameters.  

In the case of easily identifying feature correspondence, the best feature locations and relative orientations are those that result in the smallest range span length in the sidescan sonar data. 
Based on the results shown in Fig. \ref{fig:wksp_analysis_fl2ss1} and \ref{fig:wksp_analysis_fl2ss2}, feature correspondence is trivial if the feature is near the minimum range of both the forward-looking sonar and the sidescan sonar (this is to be expected, although arguably not very useful).
However, we also observe that any combination of roll, pitch, and yaw values that are all at an angle of $0$ or $\pi$ radians will produce a short length, making it easier to identify the location of the feature in both sensor outputs, regardless of the distance from the sensor. 

If the goal is to reliably retrieve the 3D location of the feature, it is best to have the longest length possible in order to achieve greater accuracy in elevation retrieval from the inverse geometric projection.
Unfortunately, our results do not indicate a clear relative pose that will guarantee the longest length. 
From the roll and pitch results in Fig. \ref{fig:wksp_analysis_fl2ss2}, the best orientation is a roll of $0$ or $\pi$ and a pitch of $\pi/2$. 
However, from the yaw and roll results in the same figure, the best roll angle is $\pi/2$. 
Rather than seeking to identify a single desirable position and orientation, we believe that a better approach would be to use the analysis to inform incremental updates, which could aid in path planning or active localization and mapping.

\subsection{Simulation of Sidescan to Forward-looking Sonar}

Our coverage for sidescan to forward-looking sonar geometry is nearly identical to the forward-looking to sidescan sonar case.
One of the two main differences is that, instead of estimating the length of the span in the range values of the sidescan sonar, we approximate the surface area of the projection in the image of the forward-looking sonar. 
The other difference is that, while the nine total simulations are nearly identical, the roles of the forward-looking and sidescan sonars are swapped.
For example, in the first simulation for forward-looking to sidescan sonar the forward-looking sonar was at a fixed distance from the feature, and in this case for the sidescan to forward-looking sonar, the sidescan sonar is what is kept at a fixed distance from the feature for the first simulation.

When considering both feature correspondence and the recovery of 3D information, the story is similar to the forward-looking to sidescan case. 
Feature correspondence is simplest when the feature is close to both the sidescan and forward-looking sonar.
However, as that is not very realistic, we also see from Fig. \ref{fig:wksp_analysis_ss2fl1} and \ref{fig:wksp_analysis_ss2fl2} that a roll angle of $\pi/2$, pitch angle of either $0$ or $\pi$, and yaw angle of either $0$ or $\pi$ is best to minimize the projection area.

Maximizing the surface area is necessary to accurately retrieve the 3D point from the inverse geometric projection, but from our results there is no clear global solution.
Instead, if given an initial relative pose, it is possible to make informed adjustments to that pose to increase the surface area. 

\section{Conclusion}

Acoustic sensing is the preferred perception modality in marine autonomous operations. Nevertheless, current state of the art acoustic-based methods lag behind optical-based approaches in many theoretical and practical areas of research. Important examples in the field of field/marine robotics in particular include 3D reconstruction and robotic localization and mapping. 
To help bridge the gap between acoustic and optical methods, we concisely lay out the projection geometry for cross-modal stereo sidescan and forward-looking sonars. 
This projection geometry, similar to epipolar geometry for optical stereo cameras, enables the projection of an observed feature from one sonar to a limited region in the observation-space of the second sonar and eases the fusion of cross-modal acoustic data. 
In addition, we present an initial analysis of the geometric projections and how they con inform hardware configurations and software algorithms to better incorporate cross-modal data for feature alignment, reconstruction, and other purposes.

Future work includes analysis of additional acoustic sensors and stereo configurations, as well as the application of our methods in real-world experiments. 

\bibliographystyle{style/IEEEtranN}
{\footnotesize
\bibliography{references/IEEEabrv,references/strings-short,references/library}}

\end{document}